\documentclass{article}  
                           
\usepackage[utf8]{inputenc}
\usepackage{hyperref}
\usepackage{graphicx}
\usepackage[margin=1in]{geometry}
\usepackage{graphicx} % Required for inserting images
\usepackage{amsmath}

\usepackage{authblk}

\title{Analyzing the Shopping Journey: Computing Shelf Browsing Visits in a Physical Retail Store}

\author[1]{Luis Yoichi Morales}
\author[2]{Francesco Zanlungo}
\author[1]{David M. Woollard}

\affil[1]{Standard AI, San Francisco CA, USA}
\affil[2]{University of Palermo, Italy}

\begin{document}
\maketitle

\section*{Abstract} 
Motivated by recent challenges in the deployment of robots into customer-facing roles within retail, this work introduces a study of customer activity in physical stores as a step toward autonomous understanding of shopper intent. We introduce an algorithm that computes shoppers' ``shelf visits'' ---  capturing their browsing behavior in the store. Shelf visits are extracted from trajectories obtained via machine vision-based 3D tracking and overhead cameras. We perform two independent calibrations of the shelf visit algorithm, using distinct sets of trajectories (consisting of 8138 and 15129 trajectories), collected in different stores and labeled by human reviewers. The calibrated models are then evaluated on trajectories held out of the calibration process both from the same store on which calibration was performed and from the other store. An analysis of the results shows that the algorithm can recognize customers' browsing activity when evaluated in an environment different from the one on which calibration was performed. We then use the model to analyze the customers' ``browsing patterns'' on a large set of trajectories and their relation to actual purchases in the stores. Finally, we discuss how shelf browsing information could be used for retail planning and in the domain of human-robot interaction scenarios.

\section{Introduction}

Despite the potential to increase customer satisfaction, improve store performance and grow sales, one of the main limiting factors to the introduction of robots in customer-facing roles is the difficulty of enabling autonomous behavior \cite{guha2022robots}. While AI promises to revolutionize the ability of robots to have open-world understanding, the general consensus is that such a revolution is far away and retailers should be conservative \cite{guha2021artificial}. Current limitations of customer-facing service robots in retail can be illustrated in Softbank’s shift from Pepper, a customer service robot designed to directly interact with customers that met with limited success, to Whiz, a robot more specifically designed for non customer-facing roles such as cleaning \cite{guha2022robots,rindfleisch2022robots}. 

A key component of missing functionality is the ability of robots to quickly ascertain shopper intent in order to directly address acute customer needs without significant assistance. While shopper intent can be considered as a complex open-world problem, we can look at parallels between shopper behaviors (and the measurement of those behaviors) in e-commerce settings for possible approaches to developing more understanding. 

Data concerning “browsing activity” are crucial in the analysis and modeling of e-commerce, but have been neglected in the performance analysis of physical retail. Nevertheless, customers of physical stores do perform parallel actions to those of online browsing. For example, shoppers can spend longer or shorter time in front of a shelf, check different products, pick up products, and then proceed to buy only some of them (if any). It is evident that this “browsing activity” correlates with customers’ purchases, and thus is extremely valuable information for both store management and Consumer Packaged Goods (CPG) manufacturers who are interested not only in their products’ sales performance, but also in how shoppers perceive and interact with products in stores.

A considerable amount of research into physical store optimization has focused on more readily available data, including the analysis of sales \cite{pesendorfer2002retail}, and at building models to optimize store layouts \cite{RePEc:spr:isochp:978-3-031-27058-1_7, RePEc:spr:isochp:978-3-031-27058-1_5} and product placement \cite{RePEc:spr:isochp:978-3-031-27058-1_8, RePEc:spr:isochp:978-3-031-27058-1_6} in order to increase revenue.

Despite these efforts, an obvious gap in physical retail analytics is in measuring the sales conversion funnel (i.e., what happens between when a shopper enters a store and when they make their purchases). We argue that, in understanding a shopper’s browsing activities in physical stores, we can start to bridge the gap in autonomy of robots by providing more targeting and meaningful support to shoppers, leading to better outcomes, both in terms of customer support and increased sales. 

In this work we introduce an algorithm that analyzes the trajectories of shoppers moving in a physical store and extracts their “shelf visits” — i.e., records of when, where and how long customers stop in front of shelves. We use the algorithm to analyze the customers’ “browsing patterns” and their relation with “buying patterns”, and discuss possible applications, limitations and future work.

\section{Related Works}
% TODO: in this section we discuss related works
%This section presents 
Related works regarding shopper behavior analysis in physical retail stores may be divided in two main  categories,  estimation of shopper paths given a shopping list and retail shelf placement optimization.

Previous works have proposed methods to estimate the path that shoppers take in a store given a shopping list and the impact of that path on purchases. For example, \cite{RePEc:spr:isochp:978-3-031-27058-1_1} proposes to estimate profit computing the average impulse purchase on shelf end caps, while \cite{flamand_2023} estimates traffic at a shelf based on its location in order to compute impulse buying. Impulse buying in particular is important to study as store revenue is highly impacted by unplanned purchases \cite{Abratt1990UnplannedBA}, so optimization of promotional product placement \cite{optimizing_sku} is a key factor is overall store performance. 

In recent years, the retail industry has become aware of the benefits of shelf layout optimization techniques based on available retail data \cite{bianch_2021}, and several studies
%(Kuhn & Sternbeck, 2013;
%Kök et al., 2015; Bianchi-Aguiar et al., 2021).
have focused on applying shelf space allocation models in practice \cite{RePEc:spr:isochp:978-3-031-27058-1_6}. Furthermore, numerical studies applying shelf space optimization have shown that cross-space elasticity has limited impact in retail profit \cite{SCHAAL2018135}. A state-of-the-art literature review of the retail shelf planning problem is given in \cite{bianchi2021retail}. 
Layout designs which improve store revenue have been previously proposed in \cite{OZGORMUS2020105562} and the relation between shelf layout and marketing effectiveness and sales is studied in 
 \cite{layout_and_marketing}. 
Also in the field of robotics some retail-oriented applications have been proposed. A work on digital twin models for retail logistics was presented in \cite{kumpel2021semantic} and robot data collection for store modeling is presented in \cite{Beetz2022}.

To the extent of our knowledge, previous works have presented neither models nor analysis of shelf browsing activity which is the focus on this work.
%computing shelf browsing and presents a discussion of its impact to product views and their impact to purchases.
%, interactions with products and their relation with purchases.
% nowadays the availability of complete trajectory data makes it possible to study the behavior of customers in the store from entrance to exit computing shelf browsing, interactions with products and their relation with purchases.
% We should include a -short- description of the store, of how the store related information (map) is stored, e.g. how a shelf information is stored, a description of data tracking process, and which information is available from data tracking (at least defining those observables that we use in the algorithm below). If this is done in mathematical notation, that notation may be used in the algorithm description.

\section{Data collection and processing}
% IN THIS SECTION LET'S DESCRIBE ONLY WHAT WE USE IN THIS WORK, NOT EVERYTHING WE STORE
\subsection{Target Stores}
\begin{figure}[!ht]
  \centering
  \includegraphics[width=0.7\linewidth]{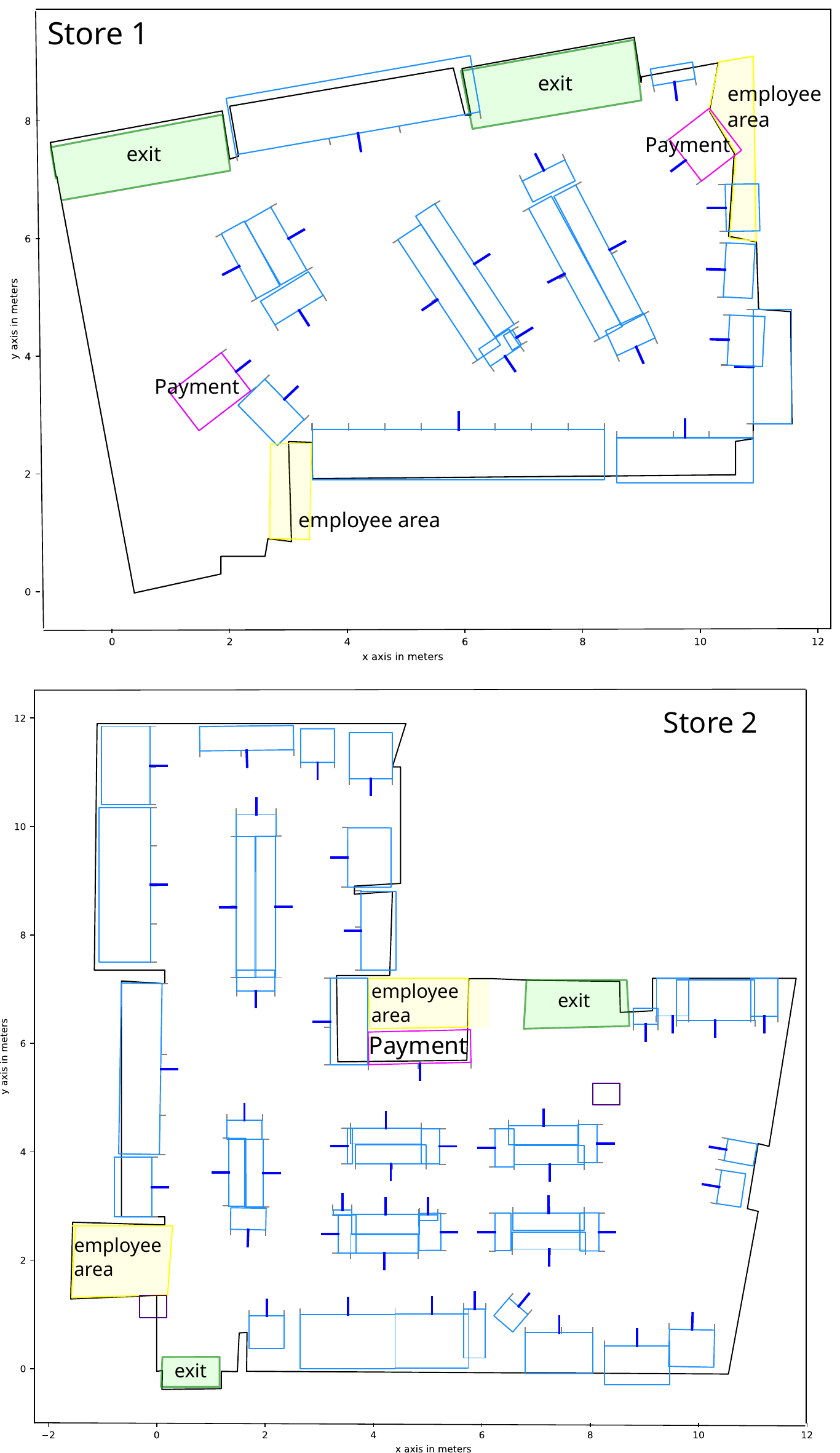}
	\caption{Layouts of store $s1$ on the top and store $s2$ on the bottom. Blue polygons show the shelves with their interactive faces with a blue line which represents the face normal.  Exits are shown in green and employee-only sections on yellow.}
  \label{fig:layouts}
\end{figure}

For this study we used data from two convenience stores (see Fig. \ref{fig:layouts}).
Store $s1$ has an area of $87.39$ m$^2$ and has $n_{s,1}=19$ shelves and $n_{e,1}=2$ exits/entrances.
Store $s2$ has an area of $109.16$ m$^2$ and has $n_{s,2}=50$ shelves and $n_{e,2}=2$ entrances/exits.

In both stores all entrances can be also used as exits, and our trajectory analysis system records if and when they are used as entrances or exits by the customers (i.e. a visit to the same entrance/exit geometrical area is recorded as a visit to an ``entrance'' at
the beginning of a trajectory tracking, and as a visit to an ``exit'' at the end).

\subsection{Store Map}
We build and use a high density colored 3D point cloud map of the store. This map is used to place cameras, compute their coverage \cite{9812042} and extract the layout of the store. The store layout contains the perimeter, entrances, exits, and shelves.  Shelves are represented as a set of 3D vertices and a normal vector that represents the  interactive shelf face (i.e., the face displaying the products). Each shelf has only 1 interactive face.
The algorithm proposed in this work takes into consideration only the projection on the 2D floor of the shelves' interactive faces and of obstacles that can obstruct the customers vision, such as the shelves' non-interacting faces.
We consider these 2D projections as segments, and we may denote the segment corresponding to the $j$th 
shelf as $\mathbf{s}_j$, with 
$j \in 1,\hdots, n_s$, and the segment corresponding to the $k$-th obstacle as $\mathbf{s}_{n_s+k}$.

% I added a description of obstacles.
\subsection{People Tracking System}
We track shopper movement using Standard AI’s Vision ML platform. 2D pose detection models, each running at 10 FPS on a calibrated set of cameras located on the ceiling of the store (similar to a typical security cameras setup) are synthesized into three-dimensional poses based on triangulation of the 2D poses from the separate cameras. We use the neck key point to track people's centroids.% and hands key points for shelf interactions.

\subsection{Data processing}
We define a trajectory as a set of time-ordered arrays including the position and body orientation of a person from the entrance to the exit of the store. The sampling rate of our tracking system is $10Hz$. We low pass filter the trajectory points before computing the velocity as described below.

Although our tracking system provides a larger amount of data (such as acceleration, hand key points, etc.) that is stored and analyzed for other purposes and future developments, the proposed algorithm uses only the following information.

The 2D vector
\begin{equation}
  \label{eq:posvector}
\mathbf{x}_i(t_k),
\end{equation}
represents the 2D projection on the store's floor of customer's $i$ (3D) centroid at the $k$th time stamp,
\begin{equation}t_k= k \Delta_t,\end{equation}
$\Delta_t=0.1$ being the tracking time step, while the angle
\begin{equation}
  \label{eq:orangle} 
\theta_i(t_k)
\end{equation}
identifies customer's $i$ body orientation (the forward normal to the 2D projection of the line connecting the shoulders, where the position of
the shoulders and the body's forward are identified using the aforementioned 3D tracking system) through the normal unit vector
\begin{equation}
\mathbf{n}_i(t_k)=(\cos(\theta_i(t_k),\sin(\theta_i(t_k)).
\end{equation}

Velocity is defined as
\begin{equation}
\mathbf{v}_i(t_k)= \frac{\mathbf{x}_i(t_{k+1})-\mathbf{x}_i(t_{k-1})}{2 \Delta_t}.
\end{equation}

Although this is a 2D vector in the following algorithm we only use its norm (speed) $v_i(t_k)$.%, and its (discrete) time derivative
%\begin{equation}
%a_i(t_k)=\frac{v_i(t_{k+1})-v_i(t_{k-1})}{2 \Delta_t},
%\end{equation}
%(which corresponds in the continuous limit to the tangential %acceleration).

% TODO Describe which tracking data are stored, and how. Describe which quantities are derived by the raw tracking data. 
\section {Shelf Stop Algorithm}
This section explains our heuristic algorithm to determine if a shelf stop happened within a trajectory. The algorithm is relatively simple, to enable calibration
on a relatively small set of trajectories, and to allow the algorithm to be used with simpler tracking systems (any tracking system providing the 2D vector $\mathbf{x}$ and
angle $\theta$ of eqs. (\ref{eq:posvector},\ref{eq:orangle}) will suffice).

The main idea is that a ``stop'' in front of a shelf $j$ is a portion of a customer' trajectory satisfying the following conditions, defined
by 3 parameters: $T_B$ (minimum browsing time), $\Delta_B$ (maximum distance to shelf) and $v_B$ (maximum browsing velocity).

%Here I modify the candidate shelf selection
More specifically, for each customer $i$ we identify a single candidate shelf (or no candidate) in the following way.
We first find all the intersections between the shelves' faces or obstacles $\mathbf{s}_k$ and the half line
defined by a positive multiple of the customer's orientation vector,
\begin{equation}\lambda \mathbf{n}_i.\end{equation}

\begin{figure}[!ht]
  \centering
  \includegraphics[width=0.7\linewidth]{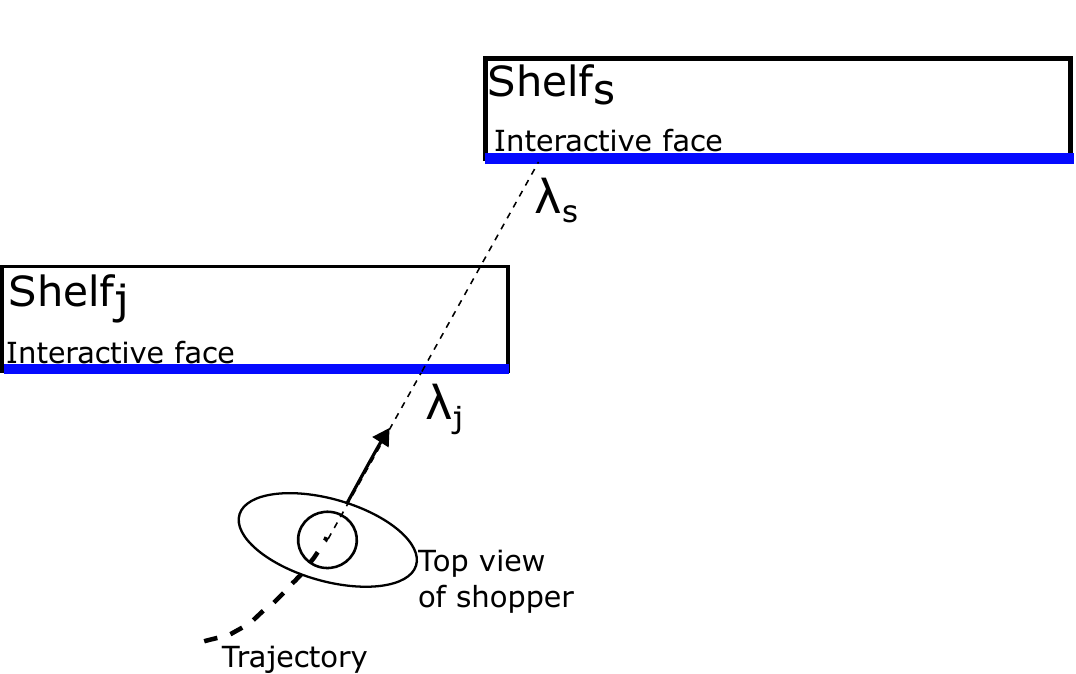}
	\caption{Top view of a shopper that may be identified as browsing shelf $j$.  $\lambda_j$ is the distance to the closest shelf, while $\lambda_s$ is the distance towards a farther shelf $s$.}
  \label{fig:shelf_counts_}
\end{figure}

We then identify each such intersection with a shelf or obstacle $l$ using
the value assumed by $\lambda$ at the intersection point,
\begin{equation}\lambda_{l}>0,\end{equation}
and look for the segment corresponding to the minimum distance,
\begin{equation}
    j=\text{argmin}_l \lambda_l.
\end{equation}

If the minimum distance corresponds to an interactive shelf face,
\begin{equation}
  j\in 1,\hdots,n_s,
\end{equation}
such shelf is identified as the candidate (otherwise, there is no candidate shelf).

This computation provides us also with the distance to the shelf,
simply defined as $\lambda_j$.

% Furthermore we consider as candidates for stops only those frame
% included in a time interval
% \begin{equation}
% t_k\in [t_s,t_f], \; t_f-t_s\geq T_B
%  \end{equation}
%  during which the candidate shelf $j$ does not change.

% Between the time stamps in the previous interval, only those satisfying the following conditions:
%  \begin{itemize}
%  \item the distance to the shelf satisfies
%   \begin{equation}
% \lambda_j(t_k)<\Delta_B,
%   \end{equation}
%  \item and the customer's velocity is smaller than a threshold value
%  \begin{equation}
% v_i(t_k)<v_B.
%  \end{equation}
%  \end{itemize}

A stop is then defined as a time interval
 \begin{equation}
t_k\in [t_s,t_f], \; t_f-t_s\geq T_B
 \end{equation}
 during which for all $k$
 \begin{itemize}
 \item the candidate shelf $j$ does not change,
 \item the distance to the shelf satisfies
 \begin{equation}
\lambda_j(t_k)\leq\Delta_B,
  \end{equation}
 \item and the customer's velocity is smaller than a threshold value
 \begin{equation}
v_i(t_k)\leq v_B.
 \end{equation}
 \end{itemize}

   For each customer $i$, shelf $j$ and time $k$ the algorithm produces a Boolean output
   \begin{equation}
   \label{eq:output}
S^i_j(t_k)
\end{equation}
assuming a value of 1 if the above conditions are satisfied, and 0 otherwise.
   \section{Algorithm Parameter Calibration}
   The values of the 3 parameters ($T_B$, $\Delta_B$ and $v_B$) are optimized through a calibration process based on human labeling.

   We built two calibration sets, the first consisting of $n_1=279$ trajectories from $s1$, and the second consisting of $n_2=270$ trajectories from $s2$. For each trajectory we produced a $2D$ video including
   position, velocity and orientation information, that human reviewers then used to identify and encode the shelf browsing behavior.

   $n_l=4$ human reviewers analyzed the full trajectory sets, and identified, based on their understanding of browsing behavior, when customer $i$ performed a stop in front
   of shelf $j$. Based on this labeling, each time stamp $k$ of customer $i$ receives a ``visit boolean'' to shelf $j$ defined by a voting system.

   Namely, defining $n^i_j(t_k)$ as the number of reviewers that identified $i$ as stopping in front of $j$ at time $k$, the visit boolean is
   \begin{equation}
V^i_j(t_k)=\begin{cases}
1, \text{ if } n^i_j(t_k)>n_l/2,\\
0, \text{ if } n^i_j(t_k)\leq n_l/2.
\end{cases}
   \end{equation}

   We may then define the number of true positives as
   \begin{equation}
    TP=\#\left(k: S^i_j(t_k)=V^i_j(t_k)=1\right),      
   \end{equation}
where $S^i_j(t_k)$ in the output of the algorithm (eq.  \ref{eq:output}); the number of false positives as
 \begin{equation}
   FP=\#\left(k: S^i_j(t_k)\neq V^i_j(t_k)=0\right),
   \end{equation}
   and the the number of false negatives as
   \begin{equation}
   FN=\#\left(k: S^i_j(t_k)\neq V^i_j(t_k)=1\right).
   \end{equation}

   We define precision as
   \begin{equation}
   P=\frac{TP}{TP+FP},
   \end{equation}
   and recall as
   \begin{equation}
   R=\frac{TP}{TP+FN}.
  \end{equation}

   The calibration process is based on finding the parameter values that maximize the $F_1$ score, defined as
   \begin{equation}
F_1=\frac{2PR}{P+R}.
   \end{equation}

      We performed two independent calibrations. The calibration process based on the $s1$ calibration set provided the following parameter values, corresponding to
   a maximum value of $F_1\approx 0.86$: $T_I=2$ s, $\Delta_I\approx 1.2$ m, $v_I\approx 0.55$ m/s.

   On the other hand, in the $s2$ calibration test a value of $F_1\approx 0.89$ was attained, corresponding to the following parameter values: $T_I=1.7$ s, $\Delta_I\approx 1.55$ m, $v_I\approx 0.48$ m/s.

   The value of $\Delta_I$ is larger on $s2$, most likely  due to the different shop geometry.
   
   \section{Algorithm Evaluation}
   % We performed two different kinds of model evaluation: same store evaluation and inter-store evaluation.

\begin{figure}[!ht]
  \centering
  \includegraphics[width=\linewidth ]{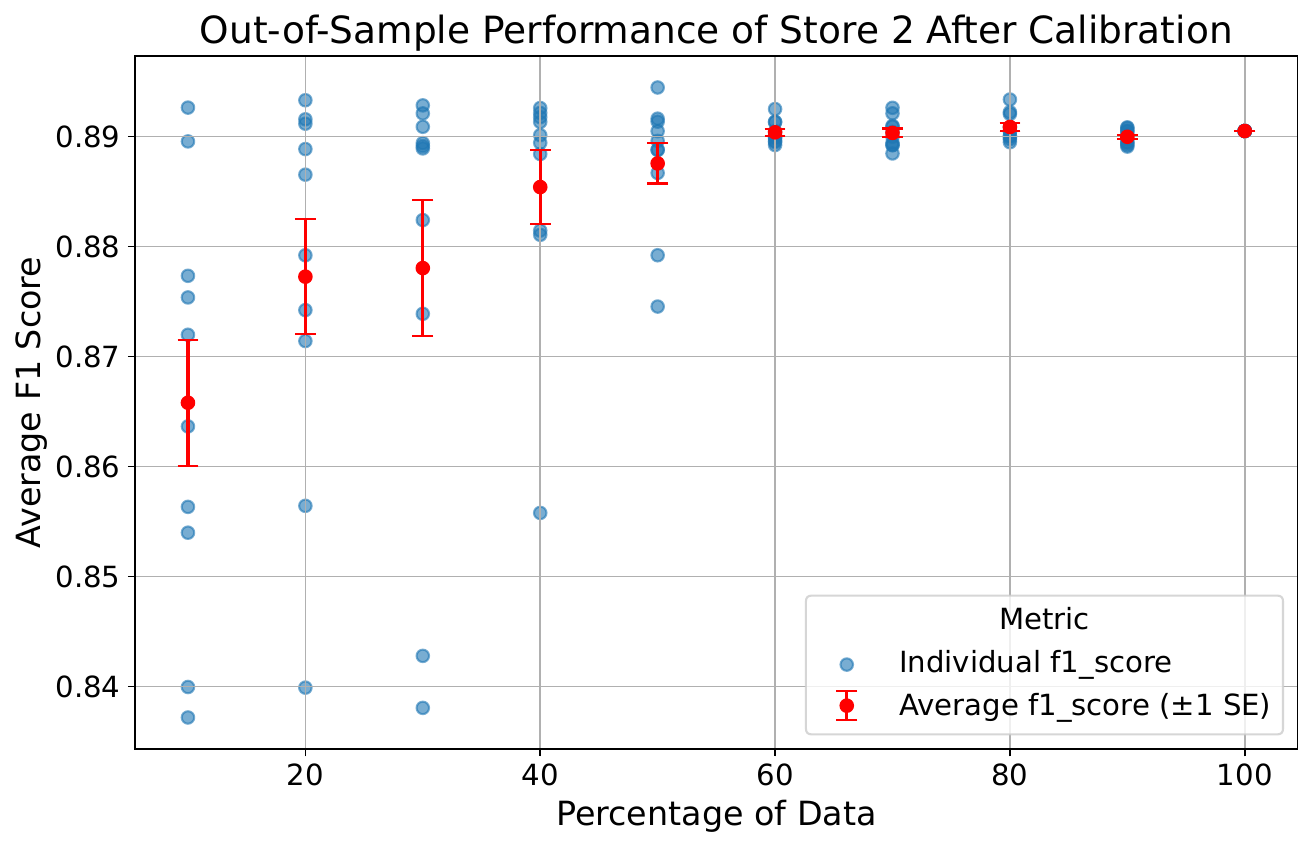}
	\caption{Same store $F_1$ score evaluation of the model. The $x$ axis shows the percentage of data used for calibration, while the $y$ axis shows the evaluation score on the remaining trajectories. The blue points correspond to 10 randomized choices of the calibration sets, the red points to the average, and the red bars to standard errors.}
  \label{fig:cal2_f1}
\end{figure}

   We performed same-store evaluation by randomly selecting a fraction $p$ of the trajectories and use it to calibrate the model, and then run the algorithm using the optimized parameters on the remaining trajectories and compute the corresponding $F_1$ score. The results concerning $s2$ are shown in Fig. \ref{fig:cal2_f1}.

   Inter-store evaluation is performed again by randomly selecting a fraction $p$ of the trajectories and using them to calibrate the model. The calibrated model is then evaluated on the full trajectory set of the other store. %The results obtained by calibrating on $s1$ and evaluating on $s2$ are shown in Fig. \ref{fig:cal1_ev2}, while the results obtained by calibrating on $s2$ and evaluating on $s1$ are shown in Fig. \ref{fig:cal2_ev1}.

   The $F_1$ score obtained by calibrating on $s1$ and evaluating on $s2$ is $0.84$ and the score obtained by calibrating on $s2$ and evaluating on $s1$ is $0.87$.
   
%   \begin{figure}[!t]
%  \centering
%  \includegraphics[width=\linewidth ]{images/calibrating store 1 testing store 2/fig4_improved_f1_score.pdf}
%	\caption{Inter-store $F_1$ score evaluation of the model. The $x$ axis shows the percentage of data from $s1$ used for calibration, while the $y$ axis shows the evaluation score on the trajectories from $s2$. The blue points correspond to 10 randomized choices of the calibration sets, the red points to the average, and the red bars to standard errors.}
%  \label{fig:cal1_ev2}
%\end{figure}

%\begin{figure}[!t]
%  \centering
%  \includegraphics[width=\linewidth ]{images/calibrating store 2 testing store 1/fig5_improved_f1_score.pdf}
%	\caption{Inter-store $F_1$ score evaluation of the model. The $x$ axis shows the percentage of data from $s2$ used for calibration, while the $y$ axis shows the evaluation score on the trajectories from $s1$. The blue points correspond to 10 randomized choices of the calibration sets, the red points to the average, and the red bars to standard errors.} 
%  \label{fig:cal2_ev1}
%\end{figure}

%    We can see that inter-store model calibration leads to an $F_1$ score $\approx 0.02$ lower than same store calibration. Using just a $p=10\%$ fraction of the data provides an acceptable score both for same and inter-store evaluation, likely owing to the simplicity of the algorithm.
    
\section{Analysis of Browsing Behavior}
We perform two kinds of analyses of browsing behavior; one on the largest set of trajectories available for $s1$ and $s2$, and one on a reduced set of $s1$ trajectories for which purchasing activity data are available.

% \subsection{Large set analysis, $s1$}
\subsection{Large set analysis, \texorpdfstring{$s1$}{s1}}
\begin{figure}[ht]
  \centering
  \includegraphics[width=\linewidth]{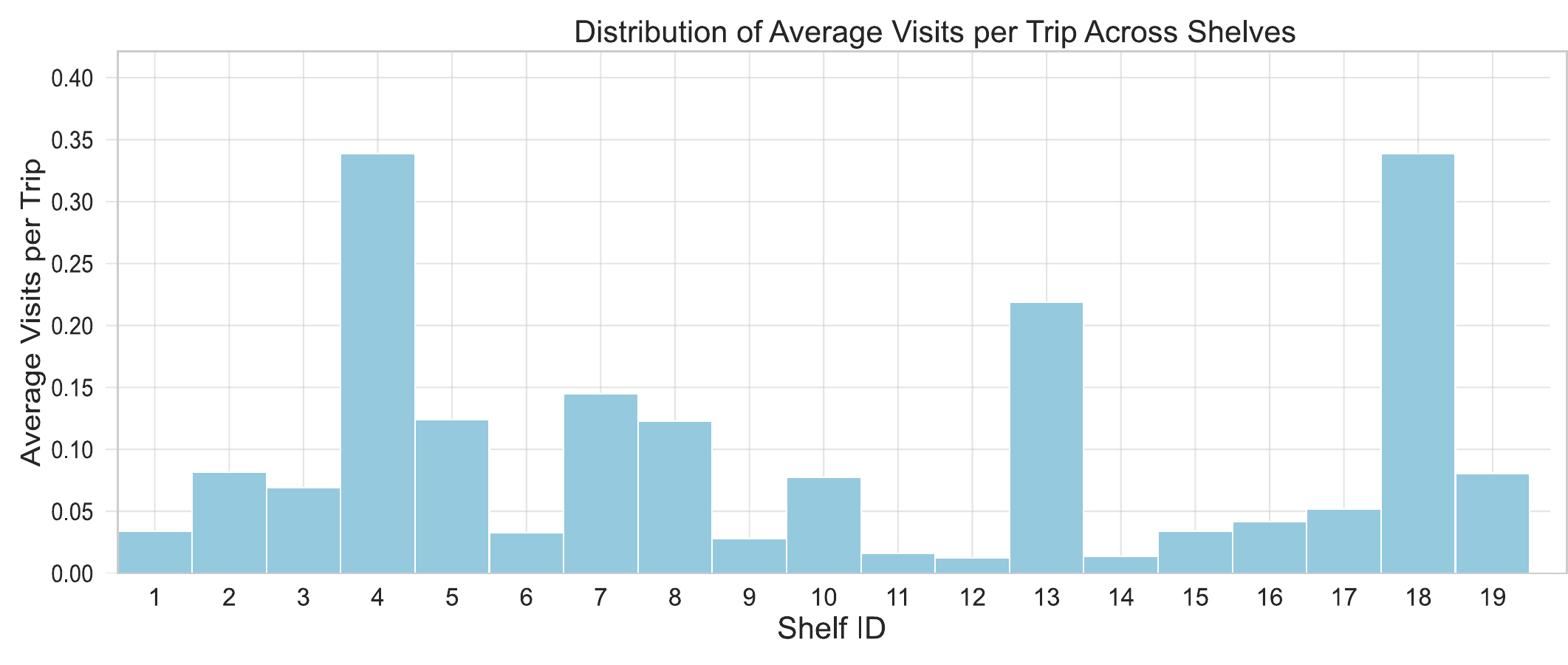}
  \caption{Distribution of average visits per trip across shelves in store $s1$.}
  \label{fig:store1_shelf_counts}
\end{figure}

We analyze $N_1=8138$ trajectories for $s1$, where $n_{s1}=19$ shelves are available. For each trajectory, we build a binary $n_{s1}$ dimensional vector, whose entries are 1 if the shopper visited the corresponding shelf, 0 otherwise. By averaging over all trajectories, we obtain an average number of visits per trip of $2.54$. See Fig. \ref{fig:store1_shelf_counts} for shelf visit counts.

% \subsection{Large set analysis, $s2$}
\subsection{\texorpdfstring{Large set analysis $s2$}{Large set analysis s2}}
\begin{figure}[ht]
  \centering
  \includegraphics[width=\linewidth]{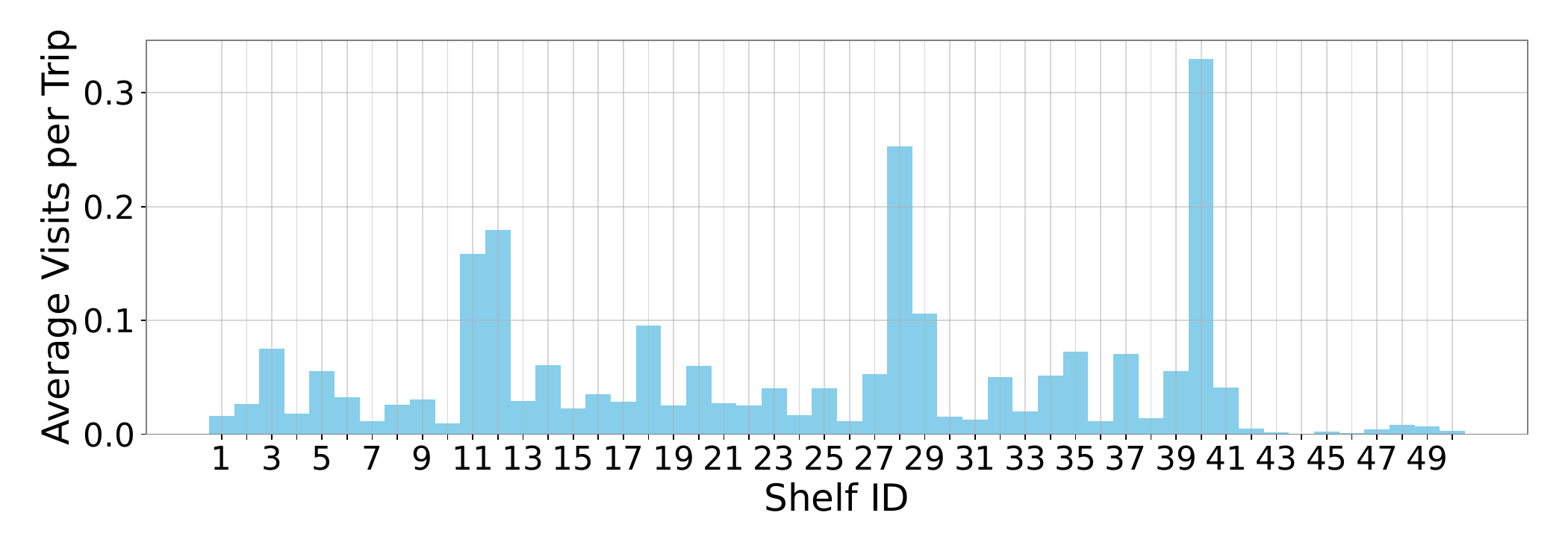}
  \caption{Distribution of average visits per trip across shelves in store $s2$.}
  \label{fig:store2_shelf_counts}
\end{figure}

We use the same method to analyze $N_2=15129$ trajectories for $s2$, where $n_{s2}=50$ shelves are available, obtaining an average number of visits per trip of $2.82$. See Fig. \ref{fig:store2_shelf_counts} for shelf visit counts.

% \subsection{Comparison to purchase activity on $s1$}
\subsection{\texorpdfstring{Comparison to purchase activity on $s1$}{Comparison to purchase activity on s1}}
For a subset of $473$ trajectories of $s1$ we have matched the trajectory to the shoppers' purchase activity (as provided by the retailers' transaction logs from their point-of-sale system). We can thus compare
the average visit vector built from these trajectories with the corresponding average number of purchases per shelf. By taking the ratio between the average purchase per shelf and the average visit per shelf, we can obtain a vector of visit/purchase conversion rates, shown in Fig 8. Although this is a very preliminary analysis, it shows the potential that our approach may have in highlighting how visits to different shelves influence purchases and ultimately provide insight into overall store performance. 

\begin{figure}[ht]
  \centering
  \includegraphics[width=\linewidth]{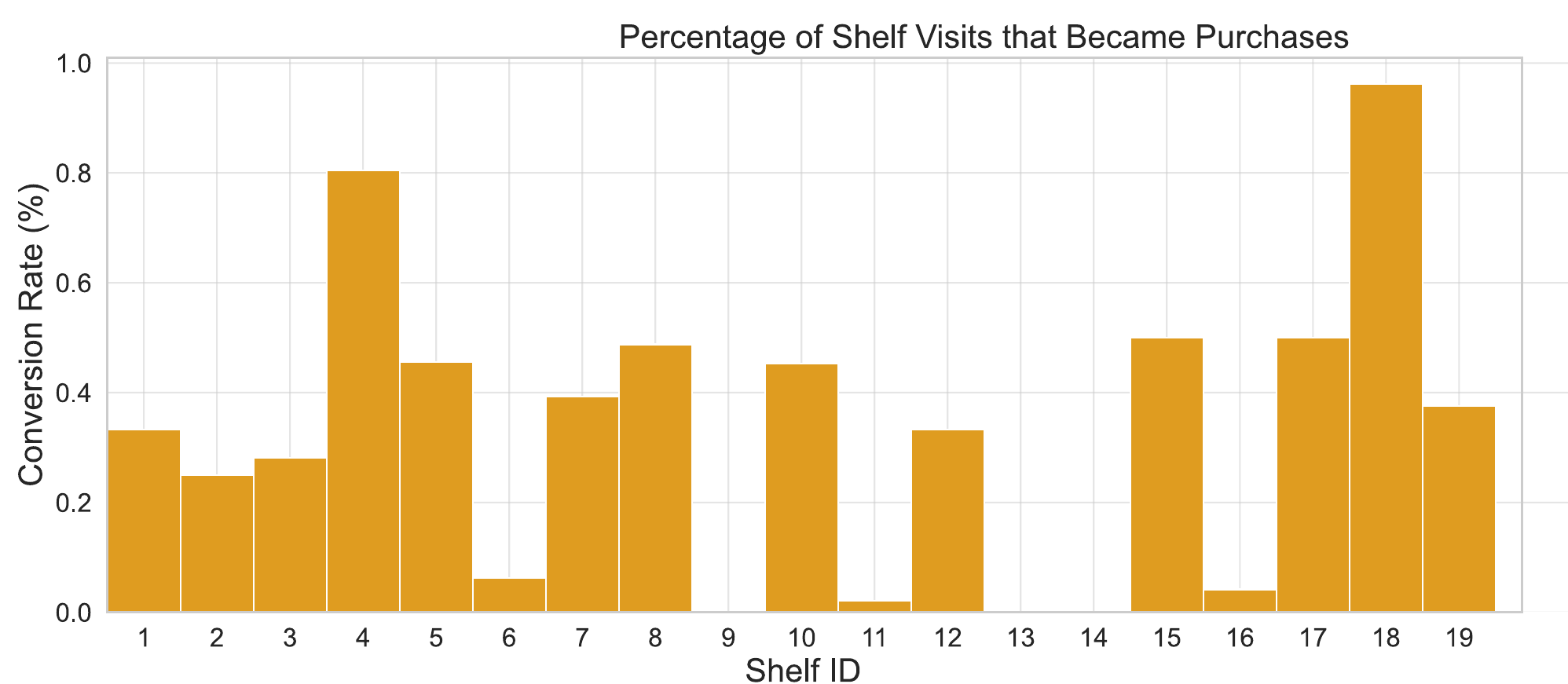}
  \caption{Percent of visits that become purchases in the subset of store $s_1$ trajectories with purchase information.}
  \label{fig:visit_to_purchase}
\end{figure}

\section{Conclusions and Future Works}
This paper presented an approach to compute shelf visits from trajectories of shoppers in physical retail stores.
The calibration of the model using human-labeled data was discussed and the evaluation of the approach using two different sets of trajectories was presented. The calibrated model was used to analyze browsing patterns and the relation between these patterns and shopper purchases.
Though this work was done in convenience stores, the same approach applies to other formats.

Future extensions of our research will involve the refinement of the model to detect more complex shopper behaviors, introduction of machine learning and other techniques to increase accuracy of browsing detection, and more targeted understanding of which products the shopper focused on during shelf browsing through analysis of gaze and head orientation. By honing in on shoppers’ intent, we can make any interventions more targeted and meaningful, including interaction with service robots whose task is to provide useful and practical help to customers \cite{shopping2}. 

Additionally, with knowledge of shopping patterns in the store, a robot system could also provide real time recommendations and advertise different product categories to shoppers based on their trajectories and previous shelf-browsing history. Similar to click histories in e-commerce, understanding shopper intent in physical stores is a significant factor in personalization of product recommendations. While this work presupposes a service robot use-case, these same capabilities (both shopper assistance and personalized recommendations) could be surfaced to shoppers through mobile applications and even in-store media displays. 

Finally, we can derive insights into how different layouts might influence shopper engagement with products. This capability will allow a data-driven approach to store layout optimization, potentially revolutionizing the way retailers think about product placement and shelf organization to maximize customer interaction and, ultimately, increase revenue.

%Autonomous checkout systems: \cite{amazongo, zippin}

\bibliographystyle{IEEEtran} % use IEEEtran.bst style
%\nocite{*}                   % list all refs in database, cited or not
%bibliography{refs}           % bib database file refs.bib
\bibliography{IEEEabrv,references}

\end{document}